\documentclass[10pt,letterpaper]{article} 
\usepackage{times}  
\usepackage{caption}
\usepackage{wrapfig}
\usepackage{enumerate}
\usepackage{graphicx}
\usepackage{multirow}
\usepackage{amsmath}
\usepackage{multirow}
\usepackage{algorithm}
\usepackage{float}
\usepackage{subfigure}
\usepackage{mathtools}
\usepackage{amssymb}
\usepackage{algpseudocode}
\usepackage{booktabs,arydshln}
\usepackage{amsthm,bm}
\usepackage{authblk}
\usepackage{xcolor}
\usepackage{url}
\usepackage{bbm}

\DeclareMathAlphabet{\fol}{OT1}{lmtt}{b}{n}

\date{}
\begin{document}
\title{Knowledge Graph Alignment using String Edit Distance}
\author{Navdeep Kaur}
\author{Gautam Kunapuli}
\author{Sriraam Natarajan}
\affil{Department of Computer Science\\
The University of Texas at Dallas}
\affil[ ]{\textit {\{navdeep.kaur,\,gautam.kunapuli,\,sriraam.natarajan\}@utdallas.edu}}
\maketitle
\section{Knowledge Graph Alignment}

Knowledge Graphs (KG) are a rich source of structured knowledge that can be leveraged to solve important AI tasks such as question answering \cite{QuestionAnsweringEmbedding2014}, relation extraction \cite{Wang2014KnowledgeGA}, recommender systems \cite{RecommenderSystem2016}. Consequently, the past decade has witnessed the development of large-scale knowledge graphs like Freebase\cite{freebase2008},  Wordnet\cite{Wordnet1995}, Yago\cite{Yago2007}, DBpedia\cite{DBpedia2014}, NELL\cite{NELL2010} that store billions of facts about the world. Typically, a knowledge graph stores knowledge in the form of triples $(h, r ,t)$ where $r$ is the relation between entity $h$ and $t$. Even though knowledge graphs are extremely large and are growing with each day, they are still incomplete with important links missing between entities. This problem of predicting  missing links between known entities is known as Knowledge Graph Completion (KBC). Over the years, embeddings based models \cite{TransE2013, HolE2016, RESCAL2011, NTN2013,  complex2016,DistMult2015} have, unarguably, become the most dominant methodology for Knowledge Graph Completion. A knowledge-graph embedding is a low-dimensional vector representation of entities and relations which are further composed by linear algebra in order to predict the missing links in a given knowledge graph.

Though highly useful in solving AI tasks, another downside of current knowledge graphs is that each of them has been developed by independent organizations by crawling facts from different sources, by utilizing different algorithms, and that sometimes, results in knowledge graphs in different languages. As a result, the knowledge embodied in these different graphs is heterogeneous and complementary \cite{ITransE2016}. This necessitates the need for integrating them in order to form one unified knowledge graph that would form a richer source of knowledge to solve AI problems more effectively. As a first step towards integrating these knowledge graphs, one needs to address the following issues, which collectively are known as $\textit{knowledge graph alignment}$: $(i)$ $\textbf{entity alignment}$ (entity resolution) that aims at finding entities in different knowledge bases being integrated which, in fact, refer to same real-world entity $(ii)$ $\textbf{triple-wise alignment}$ focuses on finding triples in two knowledge graphs that refer to the same real-world fact. For instance, even though triple ($\mathtt{m.02mjmr, \, place\_of\_birth, \, m.02hrh0\_}$) in Freebase and triple ($\mathtt{Barack\_Obama, \, birthPlace, \, Honolulu}$) in Dbpedia represent to same fact - Barack Obama was born in Honolulu - they are represented with different identities of entities and relations in two knowledge graphs.

Motivated by their success inside single knowledge graph problems, more recently, embeddings have been employed to perform knowledge graph alignment across multiple knowledge graphs. One of the primitive work along this line is Chen et al. \cite{MtransE2017}, that encodes entities and relations of two knowledge graphs into two separate embeddings space and proposes three methods of transitioning from an embedding to its counterpart in other space. Following this work, more advanced approaches for knowledge graph alignment have been proposed that can mainly be divided into three main categories: 
\begin{itemize}
\item The first set of models overcome the problem of low availability of aligned entities and aligned triples across multiple knowledge graphs. As low availability of training data can hinder the performance of model,  these works increase the size of the training data either iteratively \cite{ITransE2016}; or via bootstrapping approach \cite{BootEA2018}; or by co-training \cite{KDCoE2018} technique.
\item Another line of research is based on the idea that in addition to utilizing the knowledge in standard relation triples, there is rich semantic knowledge present in the knowledge graphs in the form of properties and text description of entities which can be harnessed to improve the performance of model \cite{JAPE2017, MultiKE2019, NAEA2019}. 
\item The third line of research is focused on designing models that overcome the limitations of translation based embeddings models \cite{NTAM2018}, as they exploit standard Graph Convolutional Networks \cite{KBAlignmentGCN2018}, their relational variants \cite{RDGCN2019, VRGCN2019} and Wasserstein GAN \cite{OTEA2019} in order to learn the embeddings of entities and relations in multiple knowledge graphs.
\end{itemize}

\subsection{Motivation}

In this work, we propose a novel $\textit{knowledge base alignment}$ technique based upon string edit distance that addresses the following limitations of the existing models:
\begin{itemize}
    \item Even though the past techniques have exploited the supplementary knowledge present in KBs in the form of text description of entities, properties of entities as attributional embeddings; none of them has exploited the rich semantic knowledge present in the $\textit{type descriptions}$ of the entities. As shown in the past \cite{TypedNELL2014, TransT2017, TKRL2016, RescalType2015}, incorporating type information into a single KB model assist in performance boost of the model. Likewise, we conjecture a performance improvement in knowledge alignment task by utilizing the type information. Further, use of type information can help the model deal with $\textit{polysemy}$ issues present in KBs.
    \item As we explain in detail in the next section, we consider multiple possible interactions between triples of two knowledge graphs by performing all possible edit distances between two triples. This is different from the linear transformation model \cite{MtransE2017} that only considers one possible way of transformation between corresponding entities/relations in two triples. Multiple transformations allow multiple ways in which two similar triples can be brought closer to each other in embedding space.
    \item Finally, all the past models have considered triple-wise alignment between triples whereas our proposed model can find similarity between relations of any arity. For instance, if our task is to perform threshold-based classification between two relations, say, $\textit{distance}(\mathtt{advisedby(william, lisa)}$, $\mathtt{coauthor(william, lisa, tom)}) < \theta$, where $\theta$ is the threshold for positive classification, then our proposed model can find the edit distance between two relations of different arity.
\end{itemize}

\section{Knowledge Alignment by String edit distance in embedding space}

We consider a multi-lingual knowledge base $\mathcal{K}$ that consists of a set $\mathcal{L}$ of languages. Specifically, we consider two ordered language pairs $(L_{1}, L_{2}) \in \mathcal{L}^2$ where each language $L_{1} = (E_{1}, R_{1}, T_{1})$  consist of set of entities $E_{1}$, relations $R_{1}$ and triples $T_{1} =  r_{1}(h_{1}, t_{1})$. Similarly, $L_{2} = (E_{2}, R_{2}, T_{2})$. We aim at finding the distance between triples $(T_{1}, T_{2}) \in (L_{1}, L_{2})$ such that the distance between aligned triples is always less than misaligned triples. Formally,
\begin{equation}
    \mathtt{dist\big(\, r_{1}(h_{1}, t_{1}), \, \, r_{2}(h_{2}, t_{2}) \, \big)} < 
    \mathtt{dist\big( \, r_{1}(h_{1}, t_{1}), \, \, r_{q}(h_{q}, t_{q}) \, \big)}
    \label{eq: editdistanceobjective}
\end{equation}
where $r_{1}(h_{1}, t_{1}) \in T_{1}$, \, $r_{2}(h_{2}, t_{2}) \in T_{2}$  and $r_{q}(h_{q}, t_{q}) \in T_{2}^{'}$. The corrupted sample set $T_{2}^{'}$ is defined as  $T_{2}^{'} = \{r_{q}(h_{2}, t_{2}) \, \vert \, \forall r_{q} \in R_{2}  \} \cup \{ r_{2}(h_{q}, t_{2})  \, \vert \, \forall h_{q} \in E_{2}   \} \cup \{ r_{2}(h_{2}, t_{q}) \, \vert \, \forall t_{q} \in E_{2} \}$ where $r_{2}(h_{2}, t_{2}) \in T_{2}$.

\subsection{String-edit distance}
The distance function of our model is inspired by the edit distance computation between a pair of strings ($\textbf{x}$, $\textbf{y}$) by memoryless stochastic transducer proposed by Ristad and Yianilos \cite{RistadStringEditDistance1998, Oncina2006}. The idea was that a transducer receives an input string $\textbf{x}$ and performs a sequence of edit operations until it reaches the terminal stage when it outputs string $\textbf{y}$. $\textit{Edit operations}$, $\delta(z)$, performed by transducer were defined as: $\delta(\mathtt{a},\mathtt{b})$: substitution of character $\mathtt{a} \in \textbf{x}$ by character $\mathtt{b} \in \textbf{y}$;
    $\delta({\mathtt{a}, \epsilon})$: deletion of character $\mathtt{a} \in \textbf{x}$; $\delta({\epsilon, \mathtt{b}}):$ insertion of character $\mathtt{b} \in \textbf{y}$. One sequence of edit operations between ($\textbf{x}$, $\textbf{y}$), called $\textit{edit sequence}$, is defined as the product of all the edit operations along the sequence. The total edit distance between pair of strings is defined as the sum of all the edit sequences $\mathtt{edq}$ :
\begin{equation}
    dist(\textbf{x}, \textbf{y}) = \sum_{\mathtt{edq}} \prod_{\delta(z) \in \, \mathtt{edq}} \delta(z)
\end{equation}
The cost of edit operations, $\delta(z)$, is a learnable cost that was optimized by EM algorithm in that model.

\subsection{String-edit operation $\delta(z)$:}
\label{sec: stringeditoperation}
Stimulated by learning of string-edit distance by Ristad and Yianilos, our goal is to compute the distance between two triples in eqn ($\ref{eq: editdistanceobjective}$) by formulating them as pair of strings. We aim at considering each aligned triple pair $(T_{1}, T_{2}) \in (L_{1}, L_{2})$ such that $T_{1} \in L_{1}$ is analogous to input string $\textbf{x}$ and $T_{2} \in L_{2}$ being analogous to output string $\textbf{y}$. Specifically, by considering triple $\mathtt{r}_{j}(\mathtt{e}_{i}, \mathtt{e}_{k})$ as string $\mathtt{r_{j}e_{i}e_{k}}$, edit distance computation between two strings can be performed by making the following assumptions:
\begin{itemize}
\item Our basic unit of edit operation is one entity $\mathtt{e}$ or one relation $\mathtt{r}$. Further, each entity or each relation are represented by low-dimensional embedding.
\item Our basic edit operation are: (a) \textit{substitution} of an entity or a relation in $T_{1} \in L_{1}$ by any another entity or relation in  $T_{2} \in L_{2}$ i.e $\delta(e_{1}, e_{2})$, $\delta(e_{1}, r_{2})$, $\delta(r_{1}, e_{2})$, $\delta(r_{1}, r_{2})$ for every $e_{1} \in E_{1}, e_{2} \in E_{2}, r_{1} \in R_{1}, r_{2} \in R_{2}$  (b) \textit{deletion} of an entity or relation present in $T_{1} \in L_{1}$ i.e. $\delta(e_{1}, \epsilon), \delta(r_{1}, \epsilon)$ for every $e_{1} \in E_{1}, r_{1} \in R_{1}$   (c) \textit{insertion} of an entity or relation present in $T_{2} \in L_{2}$ i.e. $\delta(\epsilon, e_{2}), \delta( \epsilon, r_{2})$ for every $e_{2} \in E_{2}, r_{2} \in R_{2}$. We aim to perform edit operations in embedding space.
\end{itemize}

\begin{wrapfigure}{r}{0.5\textwidth}
    \centering
    \includegraphics[scale = 0.5]{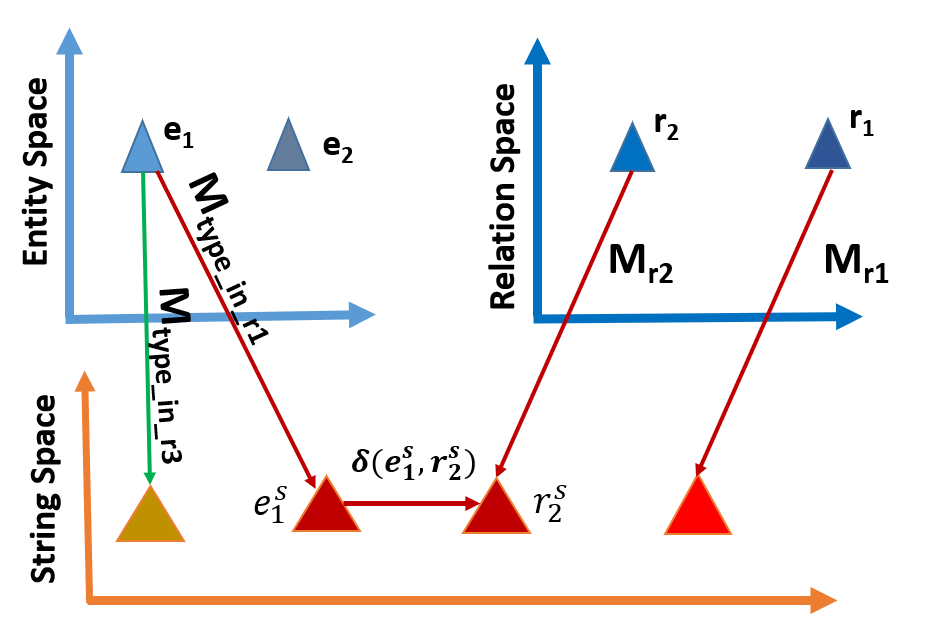}
    \caption{Knowledge graph alignment by string-edit distance in embedding space.}
    \label{fig:StringEditDistance}
    \vspace{-0.15in}
\end{wrapfigure}

As can be seen, some of the edit operations such as $\delta(e, r)$ and $\delta(r, e)$ are semantically incorrect. To overcome this, we consider three embedding spaces: entity-space, relation-space and string-space (cf. fig. $\ref{fig:StringEditDistance}$). This ensures that original entities' (or relations') information is preserved while they participate in the string-edit distance computation. Secondly, this also guarantees that entities are semantically different from relations as we locate them in separate vector space \cite{TransR2015}. 

Specifically, we model all the entities in language $L_{1}$ and $L_{2}$ to reside in $k_{e}$-dimensional embedding space, i.e. $\forall e_{1} \in E_{1}, e_{2} \in E_{2}, \mathbf{e_{1}}\in\mathbb{R}^{k_{e}}, \mathbf{e_{2}} \in \mathbb{R}^{k_{e}} $. Further, all the relations in $L_{1}$ and $L_{2}$ lie in $k_{r}$-dimensional embedding space, i.e. $\forall r_{1} \in R_{1}, r_{2} \in R_{2}, \mathbf{r_{1}} \in \mathbb{R}^{k_{r}}, \mathbf{r_{2}} \in \mathbb{R}^{k_{r}}$. In order to perform the edit operation between two triples $(T_{1}, T_{2}) \in (L_{1}, L_{2})$, their constituent entities and relations are first projected onto the $k_{s}$-dimensional string-space. For example, embedding corresponding to the triple  $\mathtt{r_{1}(h_{1}, t_{1})} \in T_{1}$ and $\mathtt{r_{2}(h_{2}, t_{2})} \in T_{2}$ in equation ($\ref{eq: editdistanceobjective}$) are projected onto string-space as follows:
\begin{equation}
    \mathbf{r_{1}^{s}} = \mathbf{r_{1}}\mathbf{M_{r_{1}}}, \, \, \, \mathbf{r_{2}^{s}} = \mathbf{r_{2}}\mathbf{M_{r_{2}}}, 
    \label{eq: projectR}
\end{equation}
\begin{equation}
    \mathbf{h_{1}^{s}} = \mathbf{h_{1}}\mathbf{M^{r_{1}}_{h_{1}-type}}, \, \, \,
    \mathbf{t_{1}^{s}} = \mathbf{t_{1}}\mathbf{M^{r_{1}}_{t_{1}-type}}, \, \, \,
    \mathbf{h_{2}^{s}} = \mathbf{h_{2}}\mathbf{M^{r_{2}}_{h_{2}-type}}, \, \, \,
    \mathbf{t_{2}^{s}} = \mathbf{t_{2}}\mathbf{M^{r_{2}}_{t_{2}-type}},
    \label{eq: projectE}
\end{equation}
where $\mathbf{r_{1}}, \, \,  \mathbf{r_{2}} \in \mathbb{R}^{k_{r}}, \, \, 
\mathbf{h_{1}}, \, \, \mathbf{h_{2}}, \, \, \mathbf{t_{1}}, \, \, \mathbf{t_{2}} \in \mathbb{R}^{k_{e}}, \, \,
\mathbf{M_{r_{1}}}, \, \mathbf{M_{r_{2}}} \in \mathbb{R}^{k_{r} \times k_{s}}, \, \, \mathbf{M^{r_{1}}_{h_{1}-type}}, \, \, \mathbf{M^{r_{1}}_{t_{1}-type}} \in \mathbb{R}^{k_{e} \times k_{s}}$ \\
$ \, \,   \mathbf{M^{r_{2}}_{h_{2}-type}}, \, \, \mathbf{M^{r_{2}}_{t_{2}-type}} \in \mathbb{R}^{k_{e} \times k_{s}}$. Also, we enforce the constraints that the embeddings and the projection matrix lie inside the unit ball i.e. $\lVert \mathbf{r^{s}} \rVert_{2} \leq 1, \, \, \lVert \mathbf{h^{s}} \rVert_{2} \leq 1, \, \, \lVert \mathbf{t^{s}} \rVert_{2} \leq 1, \, \, \lVert \mathbf{r}\mathbf{M_{r}} \rVert_{2} \leq 1, \, \,  \lVert \mathbf{e}\mathbf{M^{r}_{e-type}} \rVert_{2} \leq 1$.

The matrices $\mathbf{M_{r_{1}}}$ and $\mathbf{M_{r_{2}}}$ are the projection matrices that project the relations from the relation-space to the string-space. Similarly, $\mathbf{M^{r_{1}}_{h_{1}-type}}$ is the projection matrix that project entities from the entity-space to string-space. More specifically, projection matrix $\mathbf{M^{r_{1}}_{h_{1}-type}}$ represent the type-matrix that encodes the type of entity $h_{1}$ inside the relation $r_{1}$. The total number of type-matrices will be equal to total possible entity types in a knowledge base.

Once the entities and relations of the aligned pairs have been projected to the string-space, they are considered semantically equal. Henceforth, they represent characters of strings upon which we perform string-edit distance operations in the string-space. Consequently, aligned triples $(T_{1}, T_{2}) = \big(\mathtt{r_{1}(h_{1}, t_{1})}, \mathtt{r_{2}(h_{2}, t_{2})} \big)$ provided as training data represent transformed triple $(T_{1}, T_{2}) = \big(\mathtt{r_{1}^{s}(h_{1}^{s}, t_{1}^{s})}, \mathtt{r_{2}^{s}(h_{2}^{s}, t_{2}^{s})} \big)$ after projection. These transformed triples are modeled as string pair $(\textbf{x}, \textbf{y}) = \big(\mathtt{r_{1}^{s}h_{1}^{s}t_{1}^{s}, \, \, r_{2}^{s}h_{2}^{s}t_{2}^{s}} \big)$ in string-space, where each character of the string has its corresponding embedding, which is obtained by projection operation on entities and relations residing in their original embedding space. As a next step, we consider embeddings of characters of string $\textbf{x}$ as set $\textbf{a} = \{ \mathbf{r_{1}^{s}}, \mathbf{h_{1}^{s}}, \mathbf{t_{1}^{s}} \}$ and string $\textbf{y}$ as $\textbf{b} = \{ \mathbf{r_{2}^{s}}, \mathbf{h_{2}^{s}}, \mathbf{t_{2}^{s}} \}$ and define edit operations - substitution, deletion and insertion as follows:
\begin{itemize}
    \item \textit{substitution} operation is difference between embedding of $\textbf{a}$ and $\textbf{b}$, i.e. $\delta(\textbf{a}, \textbf{b}) = (\textbf{a} - \textbf{b}), \, \textbf{a}, \textbf{b} \in \mathbb{R}^{k_{s}}$
    \item \textit{deletion} operation $\delta(\textbf{a}, \bm{\epsilon})$ is the difference between embedding of character $\textbf{a}$ in input string $\textbf{x}$ and special null embedding $\bm{\epsilon}$:
         $\delta(\textbf{a}, \bm{\epsilon}) = (\textbf{a} - \bm{\epsilon}), \quad \textbf{a} \in \mathbb{R}^{k_{s}}$
    \item \textit{insertion} operation $\delta(\bm{\epsilon}, \textbf{b})$is the difference between special null embedding $\bm{\epsilon}$ and embedding of character $\textbf{b}$ in the output string $\textbf{y}$:       $\delta(\bm{\epsilon}, \textbf{b}) = (\bm{\epsilon} - \textbf{b}),  \quad \textbf{b} \in \mathbb{R}^{k_{s}}$

\end{itemize}

The next step after computing the edit-operation is determining the edit-sequence between string pair, which is explained in the next section.

\subsection{Edit-sequence and the Edit-distance computation}

As discussed previously, one edit-sequence is a sequence of $\textit{edit operations}$, $\delta(z)$, performed between a pair of strings $(\textbf{x}, \textbf{y})$ starting at input string $\textbf{x}$ and reaching output string $\textbf{y}$. We define one edit-sequence as an element-wise dot product of embeddings obtained as a result of edit operation, $\delta(z)$, between string pairs ($\textbf{x}, \textbf{y}$). This is followed by $\textit{L2-norm}$, in order to obtain a scalar value for one possible edit distance between ($\textbf{x}, \textbf{y}$). Formally,
\begin{equation}
    \mathtt{edq\big( r_{1}(h_{1},t_{1}), r_{2}(h_{2}, t_{2}) \big)} = \lVert \odot \big(\delta(z_{1}), \, \delta(z_{2}), \, \ldots ,\delta(z_{k}) \big) \rVert_{2}^{2} = \sum_{i=1}^{k_{s}}\big[\delta(z_{1})^{(i)}\delta(z_{2})^{(i)}  \ldots \delta(z_{k})^{(i)} \big]^{2} 
    \vspace{-0.10in}
\end{equation}
where $\delta(z_{1}), \, \delta(z_{2}), \ldots \, ,\delta(z_{k})$ are the vector obtained for each edit operation previously in the string-space. $\odot$ is the element-wise dot product of the vectors and $\delta(z_{k})^{i}$ is the $i$-th element of the vector $\delta(z_{k})$. As there can be multiple edit sequences possible between triples $(T_{1}, T_{2})$, the final distance between the pair of relation triples is defined as an average of all the edit sequences.
\begin{equation}
 \mathtt{dist\big(\, r_{1}(h_{1}, t_{1}), \, \, r_{2}(h_{2}, t_{2}) \, \big)} = \frac{1}{N}\sum_{\mathtt{edq}}\mathtt{edq\big( r_{1}(h_{1},t_{1}), \, \, r_{2}(h_{2}, t_{2}) \big)} 
 \vspace{-0.10in}
\end{equation}
where $N=\vert\mathtt{edq}\big( r_{1}(h_{1},t_{1}), \, r_{2}(h_{2}, t_{2}) \big)\vert$, number of edit sequences between triples $\mathtt{r_{1}(h_{1},t_{1}), \, r_{2}(h_{2}, t_{2})}$. To train the proposed model, we minimize margin-based ranking criteria over the aligned training pairs $(T_{1}, T_{2}) \in (L_{1}, L_{2})$:
\begin{equation}
    \mathcal{L}_{A} = \sum_{(T_{1}, T_{2})} \big[\gamma_{A} + \mathtt{dist\big(r_{1}(h_{1},t_{1}), r_{2}(h_{2}, t_{2})\big)} - \mathtt{dist\big(r_{1}(h_{1},t_{1}),r_{q}(h_{q},t_{q})\big)} \big]_{+}
\end{equation}
where $\mathtt{r_{1}(h_{1},t_{1})} \in T_{1}$ and $\mathtt{r_{2}(h_{2},t_{2})} \in T_{2}$, $[x]_{+}$ = max\{0, x\}, margin $\gamma_{A}$ is the hyperparameter. The negative example $\mathtt{r_{q}(h_{q},t_{q})}$ is obtained by corrupting positive example $\mathtt{r_{2}(h_{2}, t_{2})}$ (cf. eqn ($\ref{eq: editdistanceobjective}$)). 
\bibliographystyle{abbrv}
\bibliography{biblio}
\end{document}